\documentclass{article}
\usepackage[preprint]{preprint_style}

\usepackage{hyperref}
\usepackage{url}
\usepackage{amsmath}
\usepackage{amsmath,amssymb,bm}
\usepackage{verbatim}
\usepackage{enumitem}

\usepackage[utf8]{inputenc} 
\usepackage[T1]{fontenc}    
\usepackage{hyperref}       
\usepackage{url}            
\usepackage{booktabs}       
\usepackage{amsfonts}       
\usepackage{nicefrac}       
\usepackage{microtype}      
\usepackage{xcolor}         
\usepackage{graphicx}
\usepackage{wrapfig}
\usepackage{float}
\usepackage[ruled,vlined]{algorithm2e}
\usepackage{amsmath}
\usepackage{subcaption}
\usepackage{amsthm}

\theoremstyle{definition} 
\newtheorem{definition}{Definition}[section]

\title{Ambiguity in LLMs is a concept missing problem}

\author{%
  Zhibo Hu \\
  The University of New South Wales \\
  CSIRO Data61 \\
  Australia \\
  \texttt{zhibo.hu@student.unsw.edu.au} \\
  \And
  Chen Wang \\
  CSIRO Data61 \\
  The University of New South Wales \\
  Australia \\
  \texttt{chen.wang@data61.csiro.au} \\
  \And
  Yanfeng Shu \\
  CSIRO Data61 \\
  Australia \\
  \texttt{yanfeng.shu@data61.csiro.au} \\
  \And
  Helen Hye-Young Paik \\
  The University of New South Wales \\
  Australia \\
  \texttt{h.paik@unsw.edu.au} \\
  \And
  Liming Zhu \\
  CSIRO Data61 \\
  The University of New South Wales \\
  Australia \\
  \texttt{liming.zhu@data61.csiro.au} \\
}

\begin{document}

\maketitle

\begin{abstract}

Ambiguity in natural language is a significant obstacle for achieving accurate text to structured data mapping through large language models (LLMs), which affects the performance of tasks such as mapping text to agentic tool calling and text-to-SQL queries. Existing methods to ambiguity handling either rely on the ReACT framework to obtain correct mappings through trial and error, or on supervised fine-tuning to bias models toward specific tasks. In this paper, we adopt a different approach that characterizes representation differences of ambiguous text in the latent space and leverages these differences to identify ambiguity before mapping them to structured data. To detect sentence-level ambiguity, we focus on the relationship between ambiguous questions and their interpretations. 
Unlike distances calculated by dense embeddings, we 
introduce a new distance measure based on a path kernel over concepts.
With this measurement, we identify patterns to distinguish ambiguous from unambiguous questions. Furthermore, we propose a method for improving LLM performance on ambiguous agentic tool calling through missing concept prediction. Both achieve state-of-the-art results.

\end{abstract}

\section{Introduction}
Question answering using large language models (LLMs) often fails when user questions are ambiguous. A growing strand of work like \citep{min2020ambigqa,stelmakh2022asqa} shows that a surprising fraction of their “errors” can be traced back, not due to lack of knowledge in LLM, but to ambiguity in the user’s question itself. This ambiguity does not just mean that the question does not provide enough information, but also that the question has ambiguous semantics, i.e., multiple interpretations. 

Existing studies focus more on pragmatic or lexical ambiguity, ambiguity handling in these studies either exploits the ReACT\citep{yao2023react}
framework to produce correct mappings through trial and error, or supervised
fine tuning to guide models to produce biased mappings to improve on certain tasks\citep{saparina2025disambiguate}. \citet{kamath2024scope} attempt to use LLMs to detect ambiguity of sentences whose meaning changes with the relative scope of quantifiers, negation, or modals. They show that powerful LLMs trained on the most comprehensive datasets, such as GPT-4 sometimes default to a non-preferred semantic reading, and that success of disambiguating text varies sharply with different phrasing, which indicates disambiguation can not be easily solved by LLMs themselves. The ambiguity detection results in~\citep{saparina2024ambrosia} also confirm this observation. On the other hand, there is limited research on representational differences of ambiguous text. In this work, we study the representation of ambiguous text in the latent space and leverage the differences to
identify ambiguity.

\begin{figure*}[ht]
    \centering
    \includegraphics[width=0.65\linewidth]{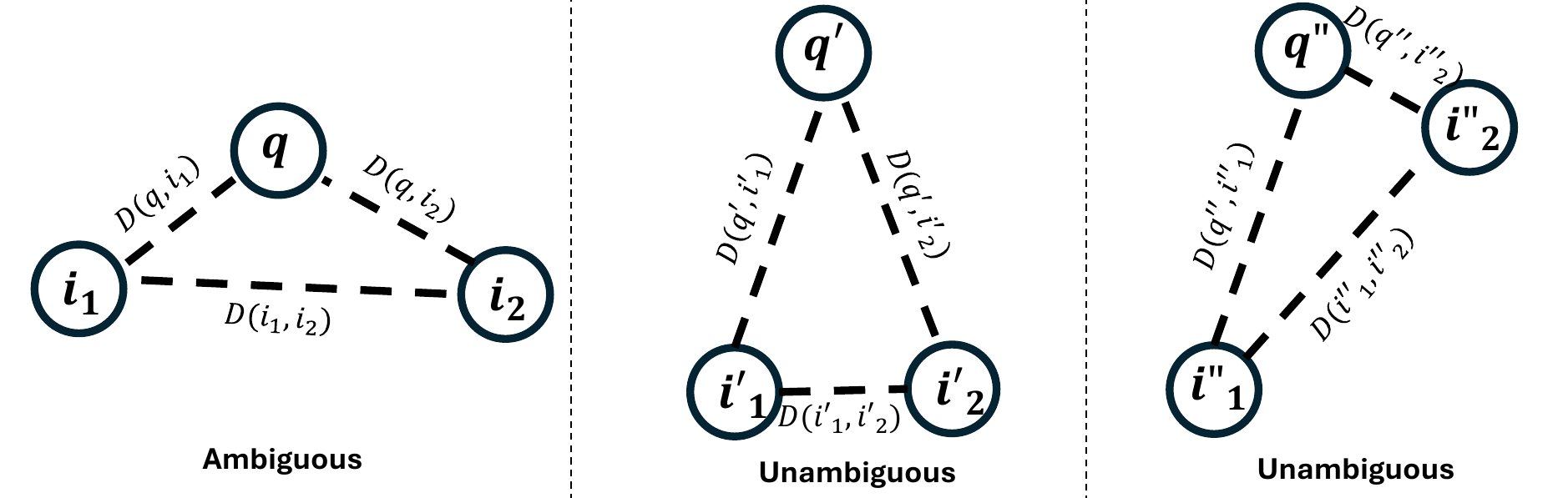}
     \caption{An example to show the difference of the distance measurement on triplets for ambiguous question ($q$, $i_1$, $i_2$) and two kind of unambiguous question ($q'$, $i'_1$, $i'_2$), ($q''$, $i''_1$, $i''_2$). For similar queries ambiguous $q$ and unambiguous $q'$, $q''$, we expect $\overline{(\mathrm{D}(q, i_1), \mathrm{D}(q, i_2), \mathrm{D}(i_1, i_2))}$ $>$ $\overline{(\mathrm{D}(q', i'_1), \mathrm{D}(q', i'_2), \mathrm{D}(i'_1, i'_2))}$ and $|\mathrm{D}(q, i_1) - \mathrm{D}(q, i_2)| \ll |\mathrm{D}(q'', i''_1) - \mathrm{D}(q'', i''_2)|$ (the overline means average).}
    \label{fig:ambigious_triplet_compare}
\end{figure*}

As an ambiguous utterance has multiple interpretations, studying the distribution of interpretations is a natural way for ambiguity detection~\citep{stengel2023zero}.  Figure \ref{fig:ambigious_triplet_compare} provides an example of the relationships between the ambiguous query $q$ and its corresponding two interpretations, denoted by $i_1$ and $i_2$. Ideally, a good distance measurement may uncover the pattern of the triplet associated with an ambiguous utterance.  
Unfortunately, current distance measurement by dense embedding vectors\citep{karpukhin2020dense} cannot give us such a measurement. The distances computed by dense vectors focus more on the semantics of individual words than on the structure of the entire sentence, which is not sensitive to the ambiguity caused by the structure of the sentence, particularly when some concepts are missing in the sentence.

In our study, we observe that the ambiguity is often associated with missing concepts in the input utterances. With the recent progress on LLM interpretability~\citep{bricken2023monosemanticity, templeton2024scaling}, the human-understandable concepts embodied in a utterance can be extracted together with their representation in the latent space through a sparse autoencoder (SAE). This inspires us to design methods to learn the concept differences of multiple interpretations of the input utterance in the latent space to identify ambiguity. We further leverage a kernel method~\citep{domingos2020every} to develop a distance metric for such latent concept comparison. 
We turn SAE into a kernel machine to measure the similarity between the concept representations of different interpretations of an ambiguous utterance. The computing is done through the integral of gradient values in a path kernel for each concept extracted by SAE. 
To make the similarity 
measurement focus on the target semantic patterns, we filter out concepts irrelevant to input utterances.  
By doing this, we successfully discover the pattern of ambiguous questions.  

Once ambiguous utterances are identified, incorporating additional information can improve 
their mapping to structured data. When the structured data requires intermediate such as SQL generation to access, users often need to be asked to clarify the utterances and provide additional information. When the structured data are finite and well-defined, e.g., tools defined within an agentic framework, this additional information can be obtained from the concepts embodied in the data.  We exploit the difference of concepts between ambiguous queries and target structured data, and design a missing concept prediction model to assist the mapping. 
We show in the experiments that our method achieves the best API calling performance on Gorilla\citep{patil2024gorilla} TensorFlow Hub bench.

In summary, our work make the following contributions:
\begin{enumerate}
    \item We observed that ambiguity arises from \textit{missing concepts} in the latent space of LLMs (Section~\ref{Ambiguity and concept missing} and \ref{Concept addition}). Using this insight, we designed a new distance measure that enhances interpretability and targets specific semantic patterns.
    \item We identify patterns to distinguish ambiguous from unambiguous questions with this measurement. 
    \item We propose a new framework to enhance the performance of LLMs in handling ambiguous agentic tool calls by predicting missing concepts.
\end{enumerate}

\section{Preliminary}

\noindent\textbf{Path Kernel.} Path kernels are used to measure how similarly a model at two data points varies during learning. Here we refer to the explanation for kernel machine from \citep{domingos2020every},
a \emph{kernel machine} predicts
\[
  y = g\!\Bigl(\sum_i a_i\,K(x,x_i) + b\Bigr),
\]
with the kernel $K$ measuring the similarity between data points.  
Gradient-descent training (learning rate $\varepsilon\!\to\!0$) implies
that the final predictor behaves like a \textbf{path kernel} machine:
\[
  \mathrm{K}_{\text{path}}(x,x') \;=\;
    \int_{c(t)} \nabla_{w} y(x)\!\cdot\!\nabla_{w} y(x') \,dt,
\]
where $c(t)$ is the parameter trajectory during training. 
The more aligned the gradients of $y$ at $x$ and $x'$, the larger the
kernel value, thus the variations of $x$ and $x'$ are more similar during training.

\noindent\textbf{Sparse autoencoder (SAE).} Neurons in modern language models often behave such that the same unit fires for several unrelated concepts. A leading hypothesis (Superposition Hypothesis) is \citep{elhage2022toy}: the model stores many more features than it has neurons by packing them into an over-complete set of directions in activation space.  
Recovering those directions is therefore a natural route to mechanistic interpretability. 
The work by Anthropic shows that a \emph{sparse auto-encoder} (SAE) trained directly on a layer’s activations can do exactly this, yielding thousands of highly interpretable, near-monosemantic concepts\citep{bricken2023monosemanticity}.

Let \(\mathbf H(x)\!\in\!\mathbb R^{d}\) denote the hidden-state (e.g.\ residual-stream) vector produced by an LLM for a token sequence \(x\).  
The goal is to learn a \emph{dictionary} \(\{\mathbf d_i\}_{i=1}^{n}\subset\mathbb R^{d}\) such that every activation can be reconstructed from a \textbf{sparse} combination of these directions:

\begin{equation} \label{eq:SAE}
\;
\mathbf H(x)\;\approx\;\mathbf b\;+\;\sum_{i=1}^{n} f_{i}\!\bigl(\mathbf H(x)\bigr)\,\mathbf d_{i}
\; 
\end{equation}

in which, \(\mathbf b\in\mathbb R^{d}\) is a learned bias that captures the mean activation. \(f_i(\mathbf H)\) is a \emph{gate} that decides whether feature~\(i\) is present; the ReLU promotes non-negativity and sparsity:
$f_i(\mathbf H)=\operatorname{ReLU}\!\bigl(\langle \mathbf w_i,\mathbf H\rangle + b_i^{\text{enc}}\bigr) $; \(\mathbf d_i\) is the \textbf{decoder} vector that take the feature back into the original space.

After training, each dictionary row \(\mathbf d_i\) corresponds to a \emph{concept}. These recovered concepts are \textbf{Sparse} (only a handful activate per token), \textbf{Linear} (they live directly in the model’s latent space), and \textbf{Monosemantic} (each gate corresponds to one dominant pattern); as a result, they can be inspected, clamped, and ablated far more easily than raw neurons.

\subsubsubsection{\textbf{Ambiguity in NLP.}} Ambiguity has been studied across various NLP tasks including machine translation~\citep{pilault-etal-2023-interactive}, natural language inference~\citep{liu-etal-2023-afraid}, question answering~\citep{kim-etal-2024-aligning, sun-etal-2023-answering}, and semantic parsing~\citep{Mu24, saparina2025disambiguate}. Recent approaches leverage LLMs to detect ambiguities by sampling multiple candidate solutions and resolving ambiguities through clarification questions or by prompting alternative interpretations. For instance, \citet{Mu24} samples multiple outputs from an LLM and examines their consistency to identify potential ambiguities. When inconsistencies are detected, the LLM is prompted to generate targeted clarification questions. However, due to the inherent biases of LLMs, the sampled solutions may lack diversity, making some ambiguities difficult to detect. To address this limitation, \citet{saparina2025disambiguate} generates an initial set of default interpretations using an LLM, which are then augmented using a specialized infilling model that requires supervised training. Our work instead examines ambiguity in the latent concept space.

\section{Methodology}

In this section, we first define the ambiguity resolution problem as a missing concept problem (\ref{Ambiguity and concept missing}). We then show the effect of adding missing concepts(\ref{Concept addition}), followed by describing our ambiguity detection method (\ref{Representation based ambiguity detection}). Finally we describe how we predict missing concepts in the context of agentic tool calling (\ref{Predict missing concepts to mitigating}).

\subsection{Ambiguity Resolution as a Missing Concept Problem}\label{Ambiguity and concept missing}

LLMs often bias towards generating one interpretation among many for an ambiguous utterance~\citep{saparina2025disambiguate}. Prompting LLMs to produce multiple interpretations and directly comparing their semantics 
do not help ambiguity detection.
To show LLMs do not produce different interpretations for an ambiguous utterance, we extract a sentence from the AMBROSIA dataset~\citep{saparina2024ambrosia} and prompt Llama-3.3-70B-Instruct\citep{meta_llama_3.3_70b} to generate two interpretations for this sentence.
\begin{figure*}[ht]
    \centering
    \includegraphics[width=0.8\linewidth]{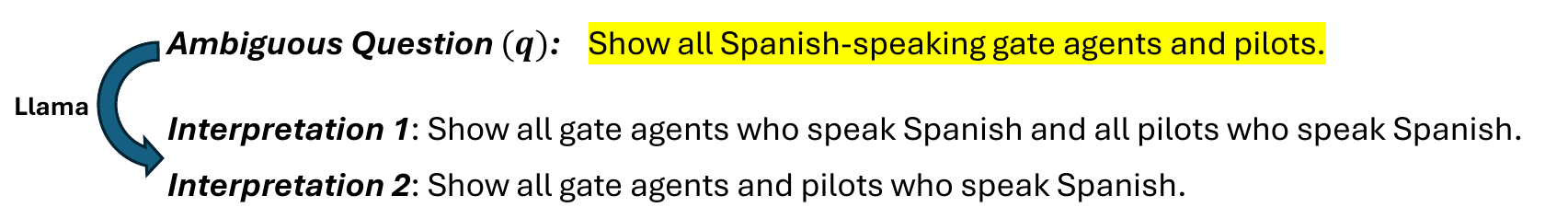}
    \label{fig:Perturbed_query_example_1}
\end{figure*}

These two interpretations in fact have the same meaning. To trigger the generation of diverse interpretations, we exploit special tokens' role in steering LLMs' responses. When we insert "[MASK]" in the original sentence, the Llama model produces two interpretations aligned with the ground-truth:
\begin{figure*}[ht]
    \centering
    \includegraphics[width=0.8\linewidth]{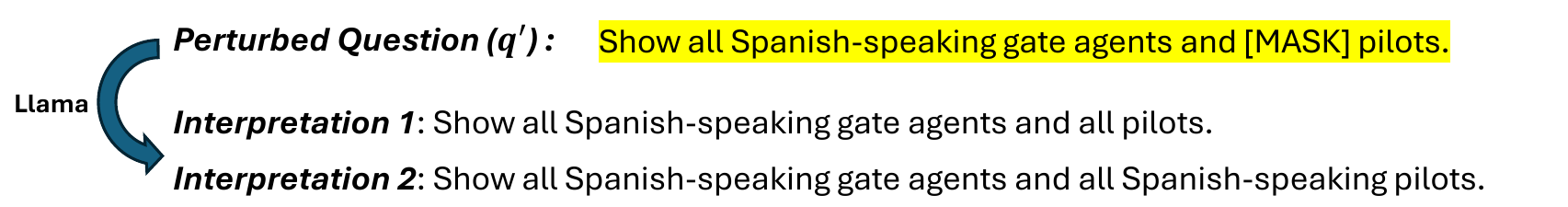}
    \label{fig:Perturbed_query_example_2}
\end{figure*}

Note that while "[MASK]" has no semantic meaning by itself, its presence in this position increases the sentence's uncertainty. 
To understand what changes are triggered by the "[MASK]" token in the concept space that make the model produce different interpretations. We use a sparse-autoencoder (SAE)\citep{goodfire_llama3.3_2025} trained on the outputs of 50 layer of this Llama model to track the new concepts after the "[MASK]" token is inserted. We get the following key concept from the SAE: 

\textit{"Start of new paragraph or point in explanatory text".}

To verify the new interpretation is indeed triggered by this concept, we individually clamp the activation value of this concept (increase it to 10) while keeping the original input sentence unchanged (without inserting the "[MASK]" token). We obtain the following interpretations:
\begin{figure}[H]
    \centering
    \includegraphics[width=0.8\linewidth]{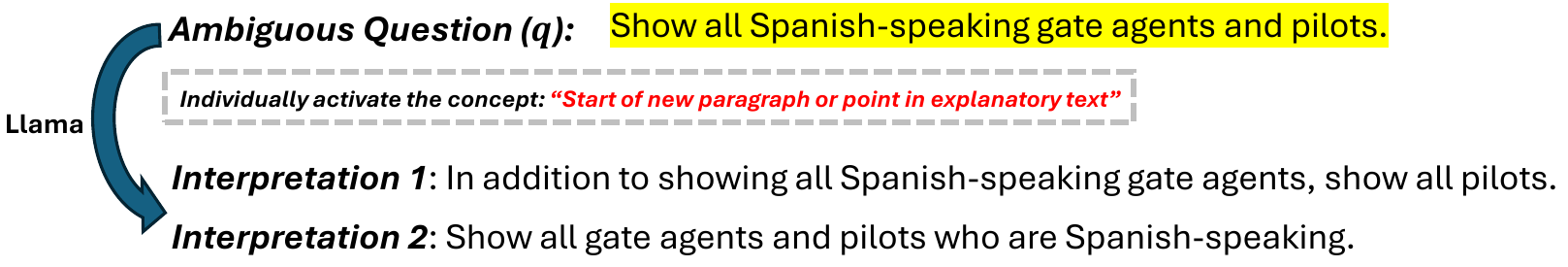}
    \label{fig:Perturbed_query_example_3}
\end{figure}

These interpretations match the ground-truth, which indicates an LLM can be steered to generate diverse interpretations for ambiguous utterances. 
We further show in the experiments that injecting examples in the prompt can effectively ``remind'' the model the missing concepts, therefore trigger the generation of diverse interpretations (see Appendix \ref{Example for concept embodied example} for examples). With the diverse interpretations of ambiguous inputs, we can then detect such ambiguity. 

However, a naive approach of using the distances of dense vectors of generated interpretations to detect ambiguity does not work well. We use the following example to elaborate this. We denote the top two interpretations of the input utterance $q$ as $i_1$ and $i_2$. By using the output of the last hidden layer of Llama-3.3-70B-Instruct for the three sentences separately, we obtain the dense vectors of the triplets $(v(q), v(i_1), v(i_2)) $:
\begin{figure}[H]
    \centering
    \includegraphics[width=0.8\linewidth]{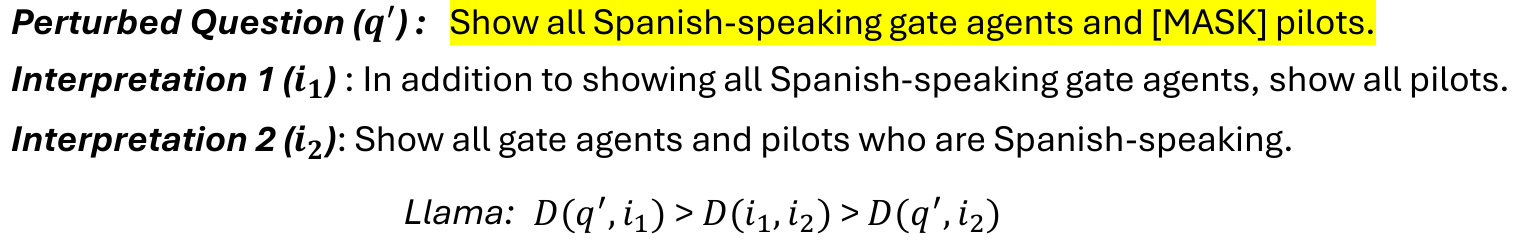}
    \label{fig:Perturbed_query_example_4}
\end{figure}

Although the "[MASK]" token activates additional concepts of the Llama model and makes it generate diverse interpretations, these additional concepts do not lead to sufficient changes in the dense vectors for ambiguity detection. The distance between $v(q')$ and $v(i_1)$ is 0.17 and the distance between $v(q')$ and $v(i_2)$ is 0.092, while the distance between $v(i_1)$ and $v(i_2)$ is 0.13. It is difficult to leverage the distance contrast in the triplet to derive a threshold to classify $q$ as ambiguous as the distance between interpretations can be arbitrarily smaller. We have done experiments with advanced embedding models and they do not have satisfactory sensitivity for distinguishing ambiguity patterns either. 

However, we notice that $q'$ and $i_1$ activated some concepts in common, which inspires us to utilize the concept differences of the triplet in the latent space to detect pattern of ambiguity. We show that such  distance measure can produce sufficient sensitivity for ambiguity detection. The distances measured using our method are as follows:
$\mathrm{D}(q', i_1) = 0.039$; $\mathrm{D}(q', i_2) = 0.027$; $\mathrm{D}(i_1, i_2) = 0.043$, meaning the distance between interpretations is larger than their distances to the query. This property produces a sensitive metric for ambiguity detection. We explain the proposed distance metric in Section~\ref{Representation based ambiguity detection}. 

\subsection{Effect of adding missing concepts into the latent space}\label{Concept addition}

\begin{wrapfigure}{r}{0.45\textwidth} 
     \includegraphics[width=\linewidth]{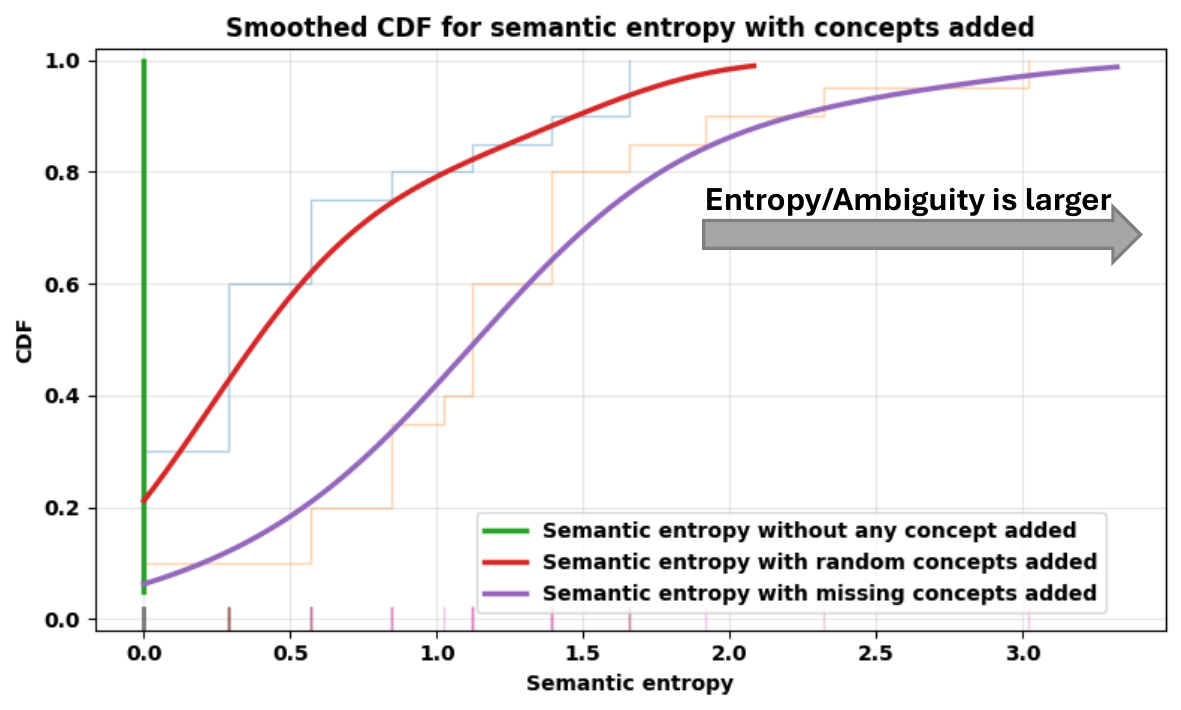}
     \vspace{-1.5\baselineskip}
     \caption{Entropy/ambiguity change with missing concepts added.}\label{fig:Entropy_with_missing_concepts_adding}
     \vspace{-0.5\baselineskip}
      
\end{wrapfigure}

To further explore the relationship between semantic ambiguity (uncertainty in LLMs) and missing concepts, we use semantic entropy\citep{kuhn2023semantic} to measure the ambiguity of query semantics (see Appendix~\ref{Measure_Semantic_Entropy_by_LLM} for algorithm details). We compute the semantic entropy  produced by LLama-3.3-70B-Instruct \citep{meta_llama_3.3_70b} on 1) 20 ambiguous questions; 2) the same 20 questions with random concepts activated; and 3) the same 20 questions with missing concepts activated.
Figure~\ref{fig:Entropy_with_missing_concepts_adding} shows that without any additional concept activated, the semantic entropy is close to zero, indicating the LLM produces a single interpretation on the input. This explains the semantic entropy of queries produced by Llama alone cannot detect ambiguity. With the missing concepts activated, the interpretations become diverse, evidenced by the increase of semantic entropy of queries. Activating random concepts also increases the semantic entropy because the noise introduced leads to diverse semantics, but the increase is less significant than that caused by adding target missing concepts.

This explains the effect of background knowledge on semantic ambiguity. For example, for this question: ``Who won the war between ethiopia and italy?'', if the LLM lacks the context of Italo-Ethiopian War, it does not know where the ambiguity is. Once the LLM retrieves the context of Italo-Ethiopian War from external sources, the concept space is enriched with the ``First War'' or the ``Second War'', which in turn increases the semantic entropy of the question.

This also explains why fine-tuning works on ambiguity resolution. Fine-tuning can be seen as learning to activate the missing concepts and therefore increase the semantic entropy of ambiguous queries. 

\subsection{Representation-based Ambiguity Detection}\label{Representation based ambiguity detection}

Our solution (see Figure~\ref{fig:Path_kernel_calculation_with_SAE}) is to use the path kernel with a sparse autoencoder (SAE) as the kernel machine for calculating distances between data points.

\noindent\textbf{SAE as Kernel Machine.} We consider SAE as $y$, the input sentences are $x$ and $x'$, their hidden states on LLM's layer where the SAE trained on are $H(x)$ and $H(x')$. Therefore, we have 
\begin{equation}
\mathrm{K}\!\bigl(x,x'\bigr)
  \;=\;
  \int_{c(t)}
    \bigl(\nabla_{\mathbf w} \text{SAE}(H(x))\bigr)\,\cdot\,
    \bigl(\nabla_{\mathbf w} \text{SAE}(H(x'))\bigr)\,dt
\end{equation}

Here, we denote the SAE on the given concept dictionary as $\mathbf{f}_{\text{SAE}}$, \noindent where \(\mathbf{f}_{\text{SAE}}(\mathbf{H(x)}) = (f_{1}(\mathbf{H(x)}),\dots,f_{N}(\mathbf{H(x)}))^{\top}\), ($f(\mathbf{H})$ in Equation~\ref{eq:SAE} ).
The \(i\)-th activation is then simply \(f_{i}(\mathbf{x})\). 

\begin{equation}
    \mathbf{f}_{\text{SAE}}(\mathbf{H(x)})
    \;=\;
    \operatorname{ReLU}\!\bigl(W_{e}(\mathbf{H(x)}-\mathbf{b}_{d})+\mathbf{b}_{e}\bigr)
    \;\in\;\mathbb{R}^{N},
\end{equation}
where $W_{e}$ is the weight matrix of the encoder and $\mathbf{b}_{d},\mathbf{b}_{e}$ are a pre-encoder and an encoder bias, respectively.  

Not all $N$ concepts are necessary for the path kernel calculation. 
To obtain the variation for input sentences $x$ and $x'$, we only need to focus on the target concepts. Accordingly, we apply a mask $M$ on the features used in gradient computation when calculating the path kernel:

\begin{align}
\mathrm{K}(x,x')
  &=
  \int_{c(t)}
      \nabla_{\mathbf w} \mathbf{f}_{\text{SAE}}^{\text{mask}}\!\bigl(\mathbf H(x)\bigr)\;\cdot\;
      \nabla_{\mathbf w} \mathbf{f}_{\text{SAE}}^{\text{mask}}\!\bigl(\mathbf H(x')\bigr)
      \,dt ,
      \label{eq:path-kernel}
\\[4pt]
\mathbf{f}_{\text{SAE}}^{\text{mask}}\!\bigl(\mathbf H(x)\bigr)
  &=
  M \circ
  \Bigl[\operatorname{ReLU}\!\bigl(W_{e}\,\bigl(\mathbf H(x)-\mathbf b_{d}\bigr)+\mathbf b_{e}\bigr)\Bigr]
      \label{eq:masked-mlp}
\end{align}
where $M$ is the concept mask (explained below) and $\circ$ is hadamard product.  

\noindent\textbf{Determining Unmasked/Target Concepts.} Semantic distances between two sentences are typically measured using the cosine similarity of their dense vectors generated by large embedding models \citep{muennighoff2022mteb}. However, this distance measurement 
is not sensitive to the semantic distinctions we want.

In interpretation generation, we use concept embodied examples (see Appendix~\ref{Example for concept embodied example}) for triggering the generation of diverse interpretations. 
To ensure our distance calculation captures the semantic meaning of sentences, we distill the concepts activated by their semantics and restrict the path kernel computation to these concepts. The distillation process involves three steps:

\begin{wrapfigure}{r}{0.45\textwidth} 
     \includegraphics[width=\linewidth]{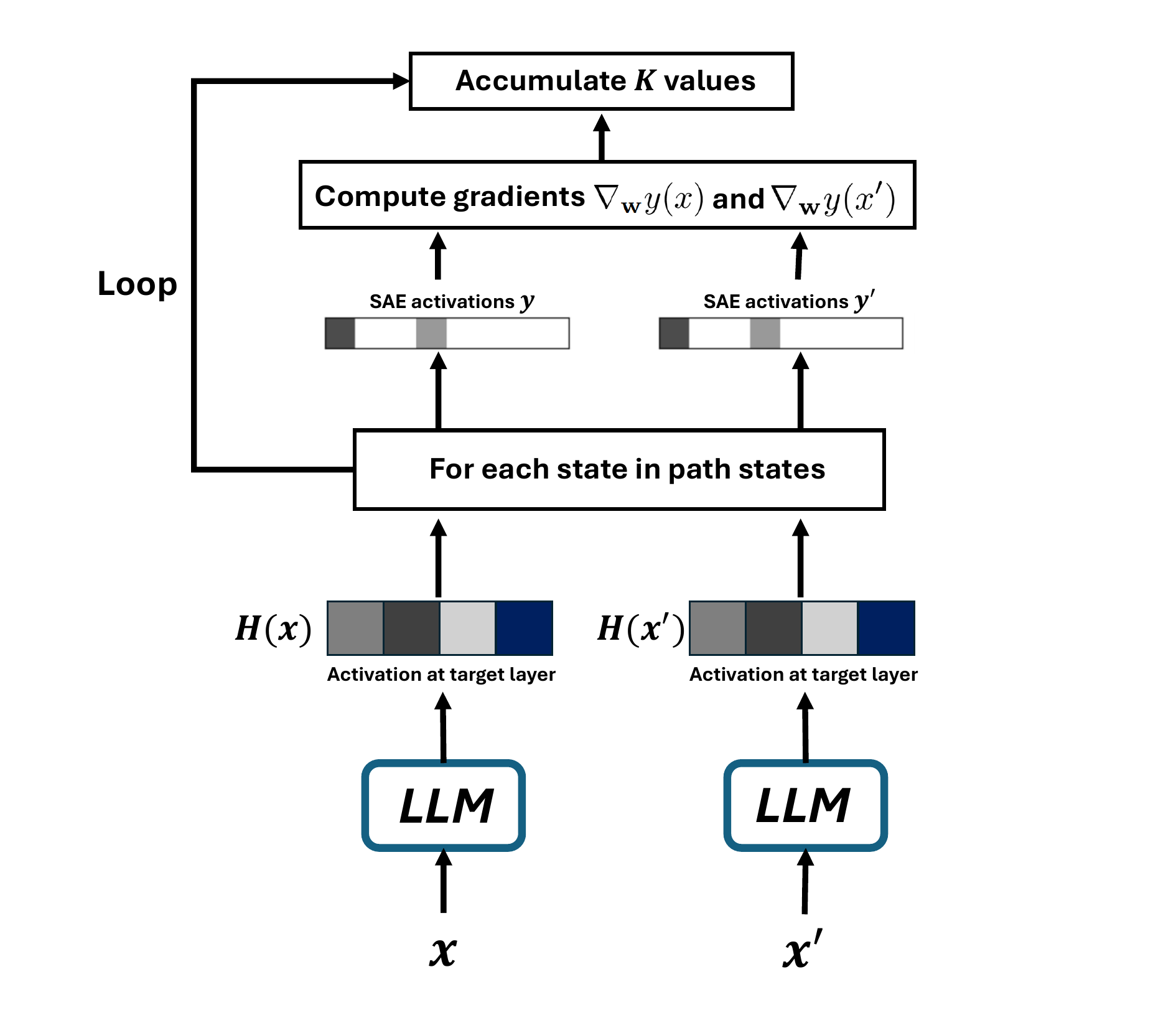}
     \vspace{-1.5\baselineskip}
     \caption{ The workflow of the path kernel calculation with SAE.}\label{fig:Path_kernel_calculation_with_SAE}
     \vspace{-0.5\baselineskip}
      
\end{wrapfigure}
\begin{enumerate}[nosep,leftmargin=12pt]
\item Collect the concepts activated by the example triplet sentences by LLM with SAE.

\item Remove the concepts activated by each individual token $t_{i}$ in the example triplet sentences from the set of concepts recorded in step 1. 

\item 
Include the remaining concepts in the mask vector $\bm{M}$, which are considered valid:

\begin{equation}
 \bm{M} = \{\mathbf{f}(\mathbf{H(x)})\} \setminus \{\mathbf{f}(\mathbf{H(t_{1}),\ldots,H(t_{n}})\} 
\end{equation}
Here $x$ is the example sentence 
and $t_{1},\ldots,t_{n}$ are the tokens in the sentence.
\end{enumerate}

\noindent\textbf{Path State Approximation.} We use a path kernel to characterize relationship of the obtained latent representations of concepts. Path states are the snapshots of a model’s parameters saved after each optimization step during training or fine-tuning. For a pre-trained SAE we usually only have the final weights, so the original series of path states cannot be reconstructed exactly. When re-training is impossible or costly, we can replace the unknown gradient-descent path with a straight-line interpolation in parameter space. 

Let 
\[
\Theta = \{\theta_k\}_{k=1}^P, \Theta^{(0)} = \{\theta_k^{(0)}\}_{k=1}^P, \Theta^* = \{\theta_k^*\}_{k=1}^P
\]
be, respectively, the parameter set, the (zero) initialization, and the final pre-trained weights.

By choosing \(n\) interpolation steps and define
$\alpha_j = \frac{j}{n-1},\quad j=0,1,\dots,n-1.$,  the \(j\)-th intermediate snapshot is then
\[
\Theta^{(j)}
=
(1 - \alpha_j)\,\Theta^{(0)}
+
\alpha_j\,\Theta^*,
\quad
\theta_k^{(j)}
=
(1 - \alpha_j)\,\theta_k^{(0)}
+
\alpha_j\,\theta_k^*
\]
Collecting them gives the full set of path states:
\[
\{\Theta^{(0)},\,\Theta^{(1)},\,\dots,\,\Theta^{(n-1)}\}.
\]

Here, $\alpha$ increases linearly from 0 to 1, forming a straight‐line path. Although this sequence does not follow the true gradient-descent dynamics, it provides a simple, deterministic path that is often adequate for estimating a path kernel.

\noindent\textbf{Distance Measurement.} The path kernel measures how similar two data points are according to the model based on their changing trajectories along the paths. To convert the (unnormalized) path kernel \(K(\cdot,\cdot)\) into a proper
distance metric between data points \(x\) and \(x'\), we apply the following two standard normalizations:

\begin{align}
\mathrm{D}_1(x,x')
   &= 1 - \frac{\mathrm{K}(x,x')}
                {\sqrt{\mathrm{K}(x,x)\,\mathrm{K}(x',x')}} ,
      \label{distance_calculation_1}\\[4pt]
\mathrm{D}_2(x,x')
   &= \sqrt{\mathrm{K}(x,x) + \mathrm{K}(x',x') - 2\,\mathrm{K}(x,x') },
      \label{distance_calculation_2}
\end{align}
We show in the experiments that both $D_1$ and $D_2$ can identify ambiguity and can be used for serving different objectives. 

\subsection{Predicting Missing Concepts to Mitigate Ambiguity} \label{Predict missing concepts to mitigating}

In Section~\ref{Ambiguity and concept missing}, we argue that the ambiguity problem arises from concepts missing in LLM's latent space, and thus distance measurements should be sensitive to this. Based on this hypothesis, we propose 
using path kernels with SAE to measure distances between questions and their interpretations. As shown in our experiments, this method reveals patterns that distinguish ambiguous questions from unambiguous ones. Motivated by this, we investigate whether ambiguity can be exploited to reduce incorrect responses and better align outputs with training data. To this end, we introduce a framework - within the context of tool calling - that retrieves data chunks by training a concept predictor on labeled data. 

\begin{figure*}[ht]
    \centering
    \includegraphics[width=0.75\linewidth]{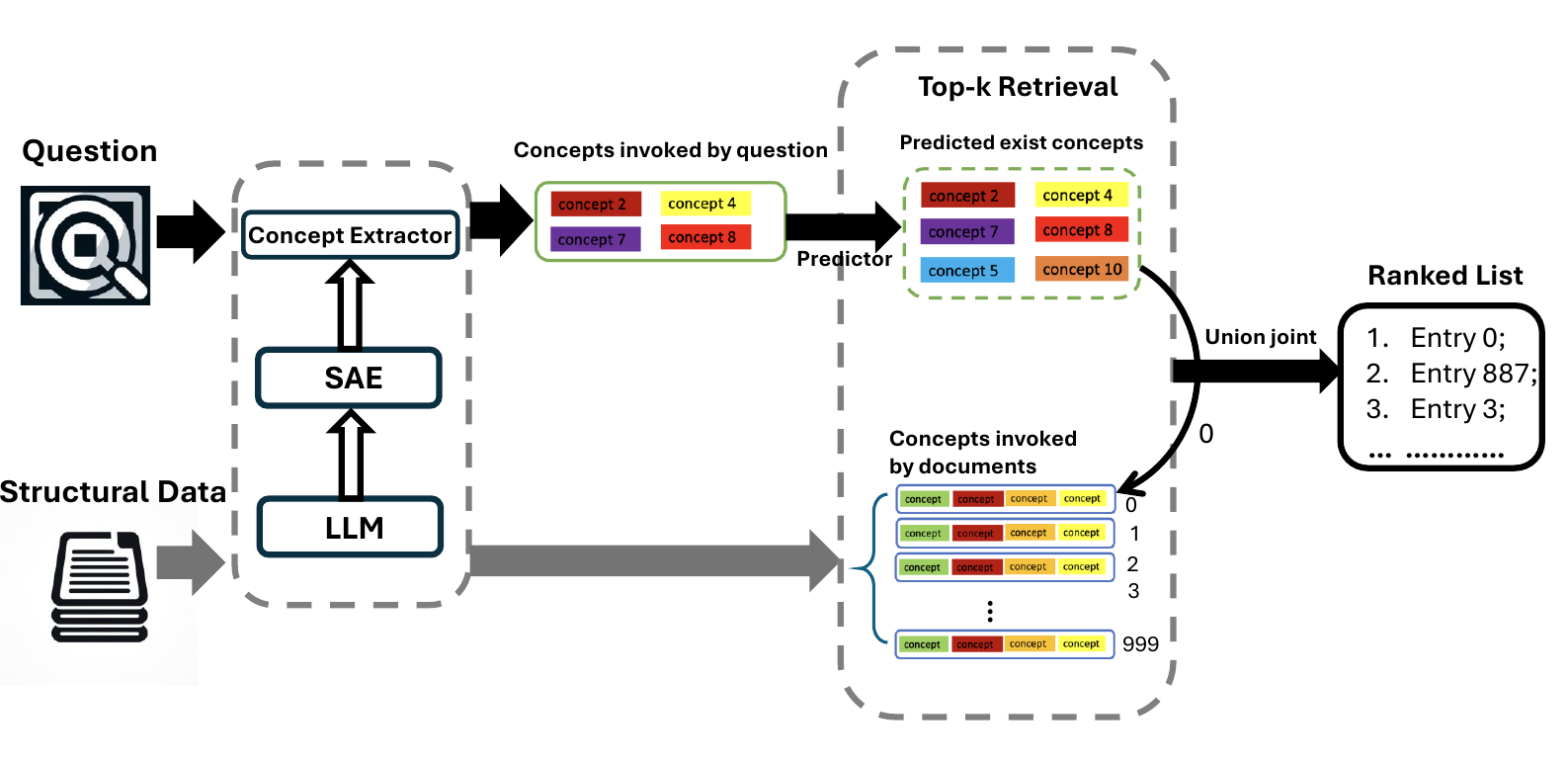}
     \caption{Tool calling framework based on missing concept prediction in ambiguous questions.}
    \label{fig:Data_chunk_Retrieval_by_SAE}
\end{figure*}

As illustrated in Figure~\ref{fig:Data_chunk_Retrieval_by_SAE}, instead of using dense embedding vectors to retrieve API calls (as in \citep{karpukhin2020dense}), we first collect the concepts activated by questions and documents on LLM
by its SAE. We then use the trained concept predictor to predict missing concepts in input questions. Finally, we rank API calls using union joint based on concept matching. Appendix~\ref{Ambiguity in agentic tool calling} presents examples of ambiguous questions in tool calling, while Appendix~\ref{Example for concepts mapping on agentic tool calling} illustrates concept matching in this context.

For efficiency, we use LightGBM \citep{ke2017lightgbm} to train the concept predictor. For each concept activated by the training data and documents, the predictor is trained to determine whether it is missing from the input question:
\begin{equation}
p\bigl(y=1 \mid x\bigr)
     \;=\;
     \sigma\!\Bigl(\sum_{t=1}^{T}\eta\,f_t(x)\Bigr)
     \;=\;
     \frac{1}{1+\exp\!\bigl(-\sum_{t=1}^{T}\eta\,f_t(x)\bigr)}
\end{equation}

\section{Experiments}
In this section, we conduct three sets of experiments:
\begin{enumerate}[nosep,leftmargin=12pt]
    \item Ambiguity detection by the proposed distance metrics: We investigate whether the distances between questions and their interpretations can distinguish ambiguous questions from unambiguous ones.
    \item Ambiguity sensitivity improvement of LLMs: We show if adding missing concepts can improve LLM's self-judgment on ambiguous questions. (see Appendix \ref{missing_concepts_adding_for_ambiguous_question})
    \item Ambiguity resolution on agentic tool calling: We investigate whether predicting missing concepts in ambiguous questions can reduce the number of incorrect responses.
\end{enumerate}

\subsection{Ambiguity Detection}

\noindent\textbf{Experiment Settings.} To evaluate the effectiveness of our method for ambiguity detection by distance differences, we conduct experiments primarily on AMBROSIA~\citep{saparina2024ambrosia}, a benchmark designed for parsing ambiguous questions into database queries across multiple domains. The benchmark consists of 1,277 ambiguous questions, each paired with human-provided unambiguous interpretations and corresponding SQL queries (2,965 in total), spanning 846 multi-table databases across 16 distinct domains. It includes three types of ambiguity—scope ambiguity, attachment ambiguity, and vagueness—and goes beyond earlier datasets that assume a single “correct” query, offering a rigorous evaluation yardstick for models that must both detect ambiguity and enumerate all valid SQL programs. 

We first prompt LLMs to generate interpretations for the ambiguous questions in the dataset. Specifically, we use LLama-3.3-70B-Instruct \citep{meta_llama_3.3_70b} for this task (see Appendix~\ref{Example for concept embodied example} for a prompting example). For each ambiguous question, we generate two interpretations, $i_1$ and $i_2$. These interpretations are then treated as unambiguous questions and are each further prompted to generate their own interpretations. As a result, for both ambiguous and unambiguous questions, we obtain two interpretations each, forming a triplet ($q$, $i_1$, $i_2$). 

Next, we compute the distances between the original question and its interpretations on AMBROSIA: $\mathrm{D}(q, i_1)$, $\mathrm{D}(q, i_2)$, and $\mathrm{D}(i_1, i_2)$. These distances are calculated using both our path kernel-based method and traditional dense vector-based methods. For comparing with both Embedding model and Generation models, we use SFR-Embedding-Mistral \citep{SFRAIResearch2024} and LLama-3.3-70B-Instruct to generate dense embeddings, with distances computed as follows:
\begin{equation}
\operatorname{D}(\mathbf{x},\mathbf{x'}) \;=\; 
1 - \frac{\mathbf{E(x)}\cdot\mathbf{E(x')}}
     {\lVert\mathbf{E(x)}\rVert\,\lVert\mathbf{E(x')}\rVert} 
\end{equation}

We further analyze the computed distances using the following two ways:
\begin{enumerate}[nosep,leftmargin=12pt]
    \item We compute the average of the three distances (by Equation~\ref{distance_calculation_1}) - $\mathrm{D}_1(q, i_1)$, $\mathrm{D}_1(q, i_2)$, $\mathrm{D}_1(i_1, i_2)$) - and plot the distribution of these mean values to show patterns.
    \item We normalize the distances (by Equation~\ref{distance_calculation_2}) using the ratios $\mathrm{D}_2(q, i_1) / \mathrm{D}_2(i_1, i_2)$ and $\mathrm{D}_2(q, i_2) / \mathrm{D}_2(i_1, i_2)$, and plot these normalized values to reveal potential patterns.
\end{enumerate}
The results from dense vector-based methods serve as baselines for comparison. 

\begin{figure*}[ht]
    \centering
    \includegraphics[width=1.\linewidth]{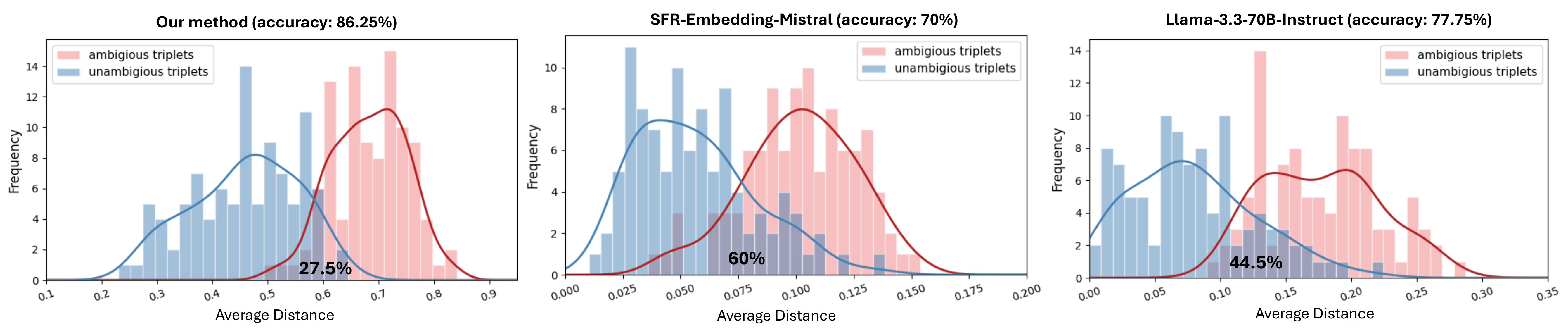}
     \caption{Distribution of average distances calculated using the path kernel method with SAE, and dense vector-based methods with SFR-Embedding-Mistral and LLama-3.3-70B-Instruct. A smaller overlapping area indicates a stronger ability to distinguish ambiguous from unambiguous questions.}
    \label{fig:distribution_images_combined}
\end{figure*}

\noindent\textbf{Results.} Figure~\ref{fig:distribution_images_combined} presents the results using our path kernel-based method (with SAE), as well as two dense vector-based methods: one using SFR-Embedding-Mistral and the other using LLama-3.3-70B-Instruct. The horizontal axis shows the average distance assigned to each sample, computed as $\overline{(\mathrm{D}_1(q, i_1), \mathrm{D}_1(q, i_2), \mathrm{D}_1(i_1, i_2))}$, for both ambiguous questions and unambiguous questions in the AMBROSIA dataset. Moving along the x-axis from left to right corresponds to increasing average distance.

The vertical axis represents the absolute frequency, i.e., the raw number of observations falling into each of the 40 equal-width histogram bins. Superimposed on the histogram bars are kernel density curves, scaled so that their peaks align with the same frequency units. This allows for a direct visual comparison between the smooth density estimates and the discrete histogram counts.

As shown in the figure, our method results in fewer overlapping samples (27.5\%) between ambiguous and unambiguous questions compared to the dense vector-based methods. Specifically, when using the x-coordinate of the intersection point of the red and blue density curves as a threshold for distinguishing ambiguous from unambiguous questions, the detection accuracies are as follows: Path kernel-based method (with SAE): 86.25\%, Dense vector method with SFR-Embedding-Mistral: 70\%, and Dense vector method with LLama-3.3-70B-Instruct: 77.75\%. As a comparison, the Zero-shot accuracy of the Llama3-70B is 46.31\%. We also visualise the distance relationship in Appendix \ref{Distance Relationship}.

\subsection{Agentic Tool Calling}

\noindent\textbf{Experiment Settings.} We evaluate our tool-calling framework (Figure~\ref{fig:Data_chunk_Retrieval_by_SAE}) on the Gorilla dataset \citep{patil2024gorilla}. This multi-faceted benchmark contains about 1.6K ML-oriented API call templates sourced from HuggingFace, TorchHub, and TensorHub. The dataset includes training and test sets, as well as API collections that support retrieval-augmented generation (RAG). 
We analyzed the API call results (including both the API calls and their domains) and found that ambiguity is a major factor contributing to performance degradation. See Appendix~\ref{Ambiguity in agentic tool calling} for an illustrative example.

We evaluate the performance of our framework on the Gorilla dataset using several baselines: the Gorilla base model (7B), the Gorilla fine-tuned model (fine-tuned on the TensorFlow Hub API dataset), and versions of fine-tuned Gorilla with BM25 and GPT-based retrievers. As the Gorilla base model is relatively small, for fair comparison, we also use a 7B model, Mistral-7B\citep{mistral7bv0.1} with its sparse-autoencoder\citep{cosgrove_mistral7b_sae16} to implement our method. Additionally, we include SFR-Embedding-Mistral \citep{SFRAIResearch2024} as a baseline~\footnote{SFR-Embedding-Mistral is ranked among the top 5 models on the MTEB leaderboard \citep{muennighoff2022mteb}.}.

To predict the missing concepts in queries, we train a LightGBM model. We then evaluate the performance of our framework on the test data by using the predicted concepts to retrieve relevant API calls from the API collections. Considering the extra computational cost introduced by the sparse autoencoder (SAE), we do not retrieve all the concepts activated by the query. Instead, we select the top 50\%, 30\%, and 20\% of the activated concepts, ranked by their activation values.

\begin{figure}[H]
    \centering
    \includegraphics[width=.65\linewidth]{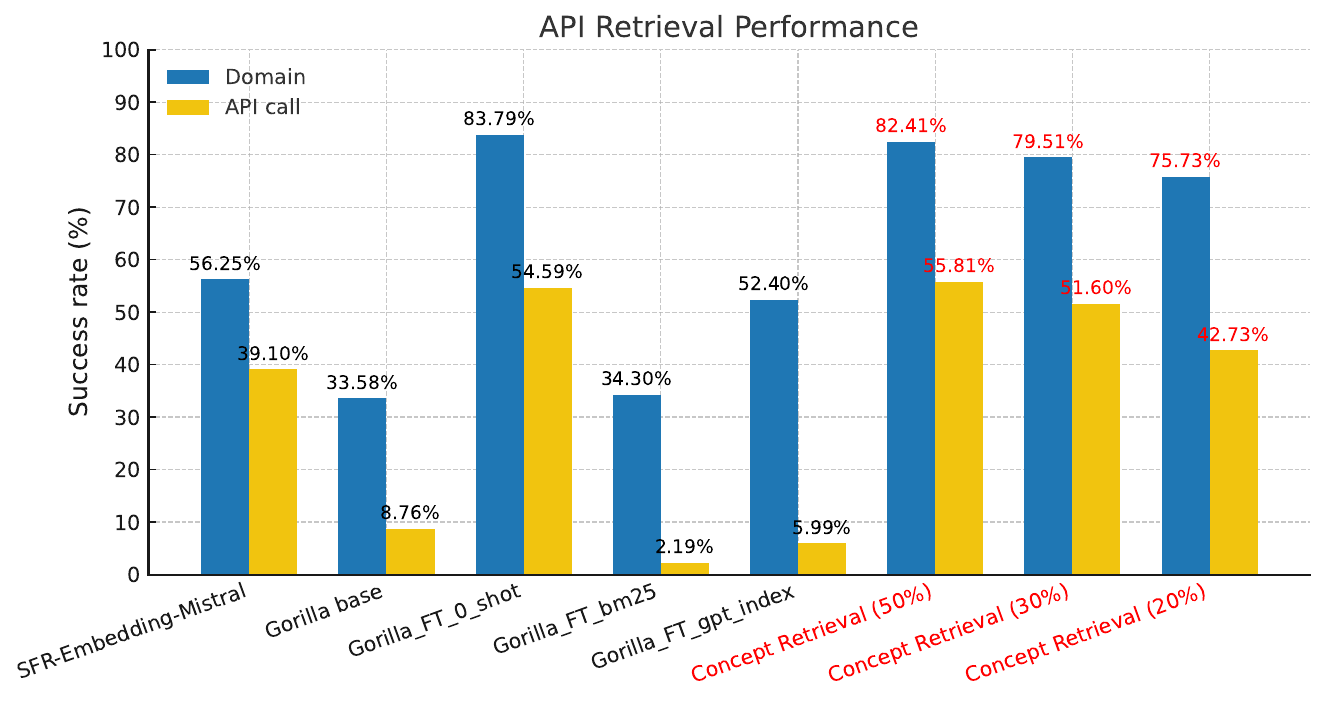}
     \caption{API bench Retrieval Results. Our methods are highlighted in red.}
    \label{fig:api_retrieval_results}
\end{figure}

\noindent\textbf{Results.} Figure~\ref{fig:api_retrieval_results} shows the performance of our concept retrieval method compared to the baselines on the Gorilla TensorFlow Hub API bench. We evaluate the accuracy of identifying the correct API domains and retrieving the correct API calls. The red highlight shows the performance of our method, demonstrating that when using the top 50\% of activated concepts, our approach achieves the highest accuracy in retrieving API calls. Accuracy of retrieving the correct domain is only slightly lower than Fine-tuned 0-shot Gorilla. We note that even when using only the top 20\% of concepts (reduce the computational cost introduced by SAE), our method still outperforms all retrieval based baselines.

\section{Conclusion}
In this paper, we designed a novel concept-based method for ambiguity resolution in LLMs. Our method distilled concepts from ambiguous utterances and their associated interpretations, inferred the pattern of their difference in the latent space and leveraged the difference for ambiguity resolution. 
We demonstrated that out method outperformed baselines on the text-to-SQL task. We also gave a new method to improve LLMs' agentic tool calling performance through missing concept prediction. The method outperformed the SOTA in APIBench. 

\bibliography{preprint}

\begin{thebibliography}{26}
\providecommand{\natexlab}[1]{#1}
\providecommand{\url}[1]{\texttt{#1}}
\expandafter\ifx\csname urlstyle\endcsname\relax
  \providecommand{\doi}[1]{doi: #1}\else
  \providecommand{\doi}{doi: \begingroup \urlstyle{rm}\Url}\fi

\bibitem[Bricken et~al.(2023)Bricken, Templeton, Batson, Chen, Jermyn, Conerly, Turner, Anil, Denison, Askell, Lasenby, Wu, Kravec, Schiefer, Maxwell, Joseph, Hatfield-Dodds, Tamkin, Nguyen, McLean, Burke, Hume, Carter, Henighan, and Olah]{bricken2023monosemanticity}
Trenton Bricken, Adly Templeton, Joshua Batson, Brian Chen, Adam Jermyn, Tom Conerly, Nick Turner, Cem Anil, Carson Denison, Amanda Askell, Robert Lasenby, Yifan Wu, Shauna Kravec, Nicholas Schiefer, Tim Maxwell, Nicholas Joseph, Zac Hatfield-Dodds, Alex Tamkin, Karina Nguyen, Brayden McLean, Josiah~E Burke, Tristan Hume, Shan Carter, Tom Henighan, and Christopher Olah.
\newblock Towards monosemanticity: Decomposing language models with dictionary learning.
\newblock \emph{Transformer Circuits Thread}, 2023.
\newblock https://transformer-circuits.pub/2023/monosemantic-features/index.html.

\bibitem[Domingos(2020)]{domingos2020every}
Pedro Domingos.
\newblock Every model learned by gradient descent is approximately a kernel machine.
\newblock \emph{arXiv preprint arXiv:2012.00152}, 2020.

\bibitem[Elhage et~al.(2022)Elhage, Hume, Olsson, Schiefer, Henighan, Kravec, Hatfield-Dodds, Lasenby, Drain, Chen, et~al.]{elhage2022toy}
Nelson Elhage, Tristan Hume, Catherine Olsson, Nicholas Schiefer, Tom Henighan, Shauna Kravec, Zac Hatfield-Dodds, Robert Lasenby, Dawn Drain, Carol Chen, et~al.
\newblock Toy models of superposition.
\newblock \emph{arXiv preprint arXiv:2209.10652}, 2022.

\bibitem[Goodfire(2025)]{goodfire_llama3.3_2025}
Goodfire.
\newblock {Llama-3.3-70B-Instruct-SAE-l50}.
\newblock \url{https://huggingface.co/Goodfire/Llama-3.3-70B-Instruct-SAE-l50}, 2025.
\newblock Accessed: 2025-05-14.

\bibitem[Kamath et~al.(2024)Kamath, Schuster, Vajjala, and Reddy]{kamath2024scope}
Gaurav Kamath, Sebastian Schuster, Sowmya Vajjala, and Siva Reddy.
\newblock Scope ambiguities in large language models.
\newblock \emph{Transactions of the Association for Computational Linguistics}, 12:\penalty0 738--754, 2024.

\bibitem[Karpukhin et~al.(2020)Karpukhin, Oguz, Min, Lewis, Wu, Edunov, Chen, and Yih]{karpukhin2020dense}
Vladimir Karpukhin, Barlas Oguz, Sewon Min, Patrick~SH Lewis, Ledell Wu, Sergey Edunov, Danqi Chen, and Wen-tau Yih.
\newblock Dense passage retrieval for open-domain question answering.
\newblock In \emph{EMNLP (1)}, pages 6769--6781, 2020.

\bibitem[Ke et~al.(2017)Ke, Meng, Finley, Wang, Chen, Ma, Ye, and Liu]{ke2017lightgbm}
Guolin Ke, Qi~Meng, Thomas Finley, Taifeng Wang, Wei Chen, Weidong Ma, Qiwei Ye, and Tie-Yan Liu.
\newblock Lightgbm: A highly efficient gradient boosting decision tree.
\newblock \emph{Advances in neural information processing systems}, 30, 2017.

\bibitem[Kim et~al.(2024)Kim, Kim, Park, Kim, Park, Yoo, Lee, and Kim]{kim-etal-2024-aligning}
Hyuhng~Joon Kim, Youna Kim, Cheonbok Park, Junyeob Kim, Choonghyun Park, Kang~Min Yoo, Sang-goo Lee, and Taeuk Kim.
\newblock Aligning language models to explicitly handle ambiguity.
\newblock In \emph{Proceedings of the 2024 Conference on Empirical Methods in Natural Language Processing}, 2024.
\newblock URL \url{https://aclanthology.org/2024.emnlp-main.119/}.

\bibitem[Kuhn et~al.(2023)Kuhn, Gal, and Farquhar]{kuhn2023semantic}
Lorenz Kuhn, Yarin Gal, and Sebastian Farquhar.
\newblock Semantic uncertainty: Linguistic invariances for uncertainty estimation in natural language generation.
\newblock \emph{arXiv preprint arXiv:2302.09664}, 2023.

\bibitem[Liu et~al.(2023)Liu, Wu, Michael, Suhr, West, Koller, Swayamdipta, Smith, and Choi]{liu-etal-2023-afraid}
Alisa Liu, Zhaofeng Wu, Julian Michael, Alane Suhr, Peter West, Alexander Koller, Swabha Swayamdipta, Noah Smith, and Yejin Choi.
\newblock We{'}re afraid language models aren{'}t modeling ambiguity.
\newblock In \emph{Proceedings of the 2023 Conference on Empirical Methods in Natural Language Processing}, 2023.
\newblock URL \url{https://aclanthology.org/2023.emnlp-main.51/}.

\bibitem[Meng et~al.(2024)Meng, Liu, Joty, Xiong, Zhou, and Yavuz]{SFRAIResearch2024}
Rui Meng, Ye~Liu, Shafiq~Rayhan Joty, Caiming Xiong, Yingbo Zhou, and Semih Yavuz.
\newblock Sfr-embedding-mistral:enhance text retrieval with transfer learning.
\newblock Salesforce AI Research Blog, 2024.
\newblock URL \url{https://www.salesforce.com/blog/sfr-embedding/}.

\bibitem[{Meta AI}(2024)]{meta_llama_3.3_70b}
{Meta AI}.
\newblock Llama-3.3-70b-instruct.
\newblock \url{https://huggingface.co/meta-llama/Llama-3.3-70B-Instruct}, December 2024.
\newblock Accessed: 2025-05-12.

\bibitem[Min et~al.(2020)Min, Michael, Hajishirzi, and Zettlemoyer]{min2020ambigqa}
Sewon Min, Julian Michael, Hannaneh Hajishirzi, and Luke Zettlemoyer.
\newblock Ambigqa: Answering ambiguous open-domain questions.
\newblock \emph{arXiv preprint arXiv:2004.10645}, 2020.

\bibitem[Mu et~al.(2024)Mu, Shi, Wang, Yu, Zhang, Wang, Liu, and Wang]{Mu24}
Fangwen Mu, Lin Shi, Song Wang, Zhuohao Yu, Binquan Zhang, ChenXue Wang, Shichao Liu, and Qing Wang.
\newblock Clarifygpt: A framework for enhancing llm-based code generation via requirements clarification.
\newblock \emph{Proc. ACM Softw. Eng.}, 1\penalty0 (FSE), 2024.
\newblock URL \url{https://doi.org/10.1145/3660810}.

\bibitem[Muennighoff et~al.(2022)Muennighoff, Tazi, Magne, and Reimers]{muennighoff2022mteb}
Niklas Muennighoff, Nouamane Tazi, Lo{\"\i}c Magne, and Nils Reimers.
\newblock Mteb: Massive text embedding benchmark.
\newblock \emph{arXiv preprint arXiv:2210.07316}, 2022.

\bibitem[Patil et~al.(2024)Patil, Zhang, Wang, and Gonzalez]{patil2024gorilla}
Shishir~G Patil, Tianjun Zhang, Xin Wang, and Joseph~E Gonzalez.
\newblock Gorilla: Large language model connected with massive apis.
\newblock \emph{Advances in Neural Information Processing Systems}, 37:\penalty0 126544--126565, 2024.

\bibitem[Pilault et~al.(2023)Pilault, Garcia, Bra{\v{z}}inskas, and Firat]{pilault-etal-2023-interactive}
Jonathan Pilault, Xavier Garcia, Arthur Bra{\v{z}}inskas, and Orhan Firat.
\newblock Interactive-chain-prompting: Ambiguity resolution for crosslingual conditional generation with interaction.
\newblock In \emph{Proceedings of the 13th International Joint Conference on Natural Language Processing and the 3rd Conference of the Asia-Pacific Chapter of the Association for Computational Linguistics (Volume 1: Long Papers)}, 2023.
\newblock URL \url{https://aclanthology.org/2023.ijcnlp-main.31/}.

\bibitem[Saparina and Lapata(2024)]{saparina2024ambrosia}
Irina Saparina and Mirella Lapata.
\newblock {AMBROSIA}: A benchmark for parsing ambiguous questions into database queries.
\newblock In \emph{The Thirty-eight Conference on Neural Information Processing Systems Datasets and Benchmarks Track}, 2024.
\newblock URL \url{https://openreview.net/forum?id=IlFk5U9cEg}.

\bibitem[Saparina and Lapata(2025)]{saparina2025disambiguate}
Irina Saparina and Mirella Lapata.
\newblock Disambiguate first parse later: Generating interpretations for ambiguity resolution in semantic parsing.
\newblock \emph{arXiv preprint arXiv:2502.18448}, 2025.

\bibitem[Stelmakh et~al.(2022)Stelmakh, Luan, Dhingra, and Chang]{stelmakh2022asqa}
Ivan Stelmakh, Yi~Luan, Bhuwan Dhingra, and Ming-Wei Chang.
\newblock Asqa: Factoid questions meet long-form answers.
\newblock \emph{arXiv preprint arXiv:2204.06092}, 2022.

\bibitem[Stengel-Eskin et~al.(2023)Stengel-Eskin, Rawlins, and Van~Durme]{stengel2023zero}
Elias Stengel-Eskin, Kyle Rawlins, and Benjamin Van~Durme.
\newblock Zero and few-shot semantic parsing with ambiguous inputs.
\newblock \emph{arXiv preprint arXiv:2306.00824}, 2023.

\bibitem[Sun et~al.(2023)Sun, Cai, Chen, Ren, Chen, de~Rijke, and Ren]{sun-etal-2023-answering}
Weiwei Sun, Hengyi Cai, Hongshen Chen, Pengjie Ren, Zhumin Chen, Maarten de~Rijke, and Zhaochun Ren.
\newblock Answering ambiguous questions via iterative prompting.
\newblock In \emph{Proceedings of the 61st Annual Meeting of the Association for Computational Linguistics (Volume 1: Long Papers)}, 2023.
\newblock URL \url{https://aclanthology.org/2023.acl-long.424/}.

\bibitem[Templeton et~al.(2024)Templeton, Conerly, Marcus, Lindsey, Bricken, Chen, Pearce, Citro, Ameisen, Jones, Cunningham, Turner, McDougall, MacDiarmid, Freeman, Sumers, Rees, Batson, Jermyn, Carter, Olah, and Henighan]{templeton2024scaling}
Adly Templeton, Tom Conerly, Jonathan Marcus, Jack Lindsey, Trenton Bricken, Brian Chen, Adam Pearce, Craig Citro, Emmanuel Ameisen, Andy Jones, Hoagy Cunningham, Nicholas~L Turner, Callum McDougall, Monte MacDiarmid, C.~Daniel Freeman, Theodore~R. Sumers, Edward Rees, Joshua Batson, Adam Jermyn, Shan Carter, Chris Olah, and Tom Henighan.
\newblock Scaling monosemanticity: Extracting interpretable features from claude 3 sonnet.
\newblock \emph{Transformer Circuits Thread}, 2024.
\newblock URL \url{https://transformer-circuits.pub/2024/scaling-monosemanticity/index.html}.

\bibitem[{The Mistral AI Team}()]{mistral7bv0.1}
{The Mistral AI Team}.
\newblock {Mistral-7B-v0.1}.
\newblock \url{https://huggingface.co/mistralai/Mistral-7B-v0.1}.
\newblock Accessed: 12 May 2025.

\bibitem[{Tyler Cosgrove}()]{cosgrove_mistral7b_sae16}
{Tyler Cosgrove}.
\newblock {Mistral-7B-sparse-autoencoder-layer16}.
\newblock \url{https://huggingface.co/tylercosgrove/mistral-7b-sparse-autoencoder-layer16}.
\newblock Accessed: 12 May 2025.

\bibitem[Yao et~al.(2023)Yao, Zhao, Yu, Du, Shafran, Narasimhan, and Cao]{yao2023react}
Shunyu Yao, Jeffrey Zhao, Dian Yu, Nan Du, Izhak Shafran, Karthik Narasimhan, and Yuan Cao.
\newblock React: Synergizing reasoning and acting in language models.
\newblock In \emph{International Conference on Learning Representations (ICLR)}, 2023.

\end{thebibliography}
\bibliographystyle{plainnat}

\clearpage

\appendix
\section{Algorithm: Measuring Semantic Entropy by LLM}\label{Measure_Semantic_Entropy_by_LLM}

\begin{definition}[Semantic Entropy]
Given a prompt $x$, let $\mathcal{S}$ be the set of possible sequences (texts) and
let $\mathcal{C}$ be a partition of $\mathcal{S}$ into semantic equivalence
classes (meanings) $c \in \mathcal{C}$. The language model induces a
distribution $p(s\mid x)$ over sequences $s \in \mathcal{S}$, which pushes
forward to a distribution over meanings,
\[
p(c\mid x) \;=\; \sum_{s \in c} p(s\mid x).
\]
The \emph{semantic entropy} of the model at $x$ is the Shannon entropy of this
meaning distribution:
\begin{equation}
\label{eq:se}
\mathrm{SE}(x)
\;=\;
- \sum_{c \in \mathcal{C}} p(c\mid x)\,\log p(c\mid x)
\;=\;
- \sum_{c \in \mathcal{C}}
\Bigg(\sum_{s \in c} p(s\mid x)\Bigg)
\log \Bigg(\sum_{s \in c} p(s\mid x)\Bigg).
\end{equation}
\end{definition}

At a high level the semantic entropy estimation involves three steps (Algorithm~\ref{alg:semantic_entropy}):

\begin{enumerate}
\item \textbf{Generation.}
Given a prompt $x$, sample $N$ sequences
$s_1,\dots,s_N \sim p(\cdot\mid x)$ from the LLM.

\item \textbf{Clustering.}
Embed each sequence $z_i=\mathcal{E}(s_i)$ and group the embeddings into
semantic equivalence classes 
$C=\{C_1,\dots,C_{|C|}\}$ via a clustering method $\mathcal{C}$ (cosine + agglomerative with threshold).
Let $\ell_i$ denote the class label of $s_i$ and $C_k=\{i:\ell_i=k\}$.
\item \textbf{Entropy estimation.}
Estimate the class masses $p(C_k\mid x)$ and compute entropy over meanings.

\end{enumerate}

\begin{algorithm}[H]
\caption{Measuring Semantic Entropy (Monte-Carlo class-average)}
\label{alg:semantic_entropy}
\KwIn{Prompt $x$; num of samples $N$; generator $\mathcal{G}$; embedder $\mathcal{E}$; clustering procedure $\mathcal{C}$ }
\KwOut{semantic entropy $H_{\mathrm{sem}}$, cluster probs $\{p_k\}$, labels $\{\ell_i\}$}

\tcp{1) Generate $N$ samples and (optional) scores}
\For{$i \leftarrow 1$ \KwTo $N$}{
  $y_i, s_i \leftarrow \mathcal{G}(x)$ \tcp*{$y_i$: text, $s_i$: (avg) sequence log-prob}
}

\tcp{2) Embed and cluster by meaning}
$Z_i \leftarrow \mathcal{E}(y_i)$ for $i=1..N$\;
$\ell_1,\dots,\ell_N \leftarrow \mathcal{C}(Z_1,\dots,Z_N)$ \tcp*{e.g., cosine + agglomerative}

\tcp{3) Estimate cluster probabilities}
Let $K$ be the number of distinct clusters among $\{\ell_i\}$\;
\If{using counts}{
  $n_k \leftarrow \sum_{i=1}^{N} \mathbf{1}[\ell_i = k]$; \quad
  $p_k \leftarrow \frac{n_k}{N}$ for $k=1..K$\;
}
\Else{\tcp{probability-weighted}
  \tcp{Stabilize weights with log-sum-exp}
  $m \leftarrow \max_i s_i$; \quad $\tilde{w}_i \leftarrow \exp(s_i - m)$; \quad
  $w_i \leftarrow \frac{\tilde{w}_i}{\sum_{j=1}^{N} \tilde{w}_j}$\;
  $p_k \leftarrow \sum_{i=1}^{N} w_i \cdot \mathbf{1}[\ell_i = k]$ for $k=1..K$\;
}

\tcp{4) Ensure numerical safety (clip and renormalize)}
$\varepsilon \leftarrow 10^{-12}$\;
$p_k \leftarrow \max(p_k, \varepsilon)$; \quad
$p_k \leftarrow \frac{p_k}{\sum_{j=1}^{K} p_j}$ for $k=1..K$\;

\tcp{5) Calculate semantic entropy}
$H_{\mathrm{sem}} \leftarrow - \sum_{k=1}^{K} p_k \log_b p_k$ \quad (default $b=2$ for bits)\;

\Return{$H_{\mathrm{sem}}$, $\{p_k\}$, $\{\ell_i\}$}\;
\end{algorithm}

\section{Ambiguity Detection: Prompt examples and Additional Results}\label{appnedix_more_experiments}

\clearpage

\subsection{Few-shot Examples used in Prompts to Inject Concepts}\label{Example for concept embodied example}

\begin{figure*}[h!]
    \centering
    \includegraphics[width=0.85\linewidth]{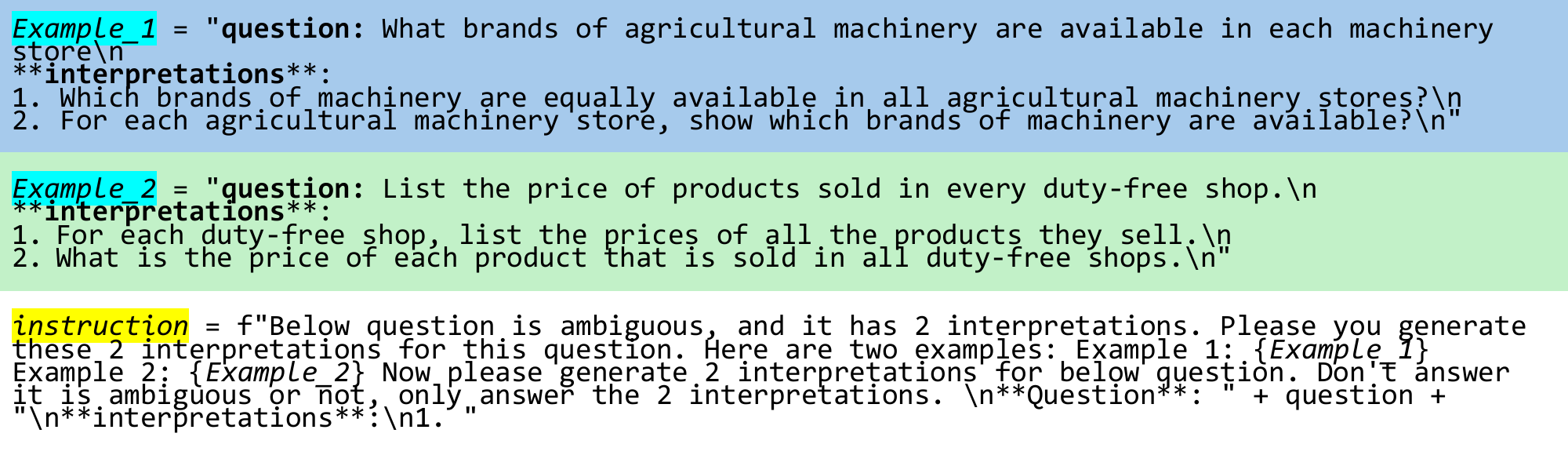}
     \caption{An example to prompt LLM to generate diverse interpretations.}
    \label{fig:Example for concept embodied example}
\end{figure*}

Here we provide an example 
showing how concept-embodied examples can guide an LLM to generate diverse interpretations (Figure~\ref{fig:Example for concept embodied example}).

\subsection{Distance Relationship}\label{Distance Relationship}

\begin{figure}[h!]
    \centering
    \includegraphics[width=0.85\linewidth]{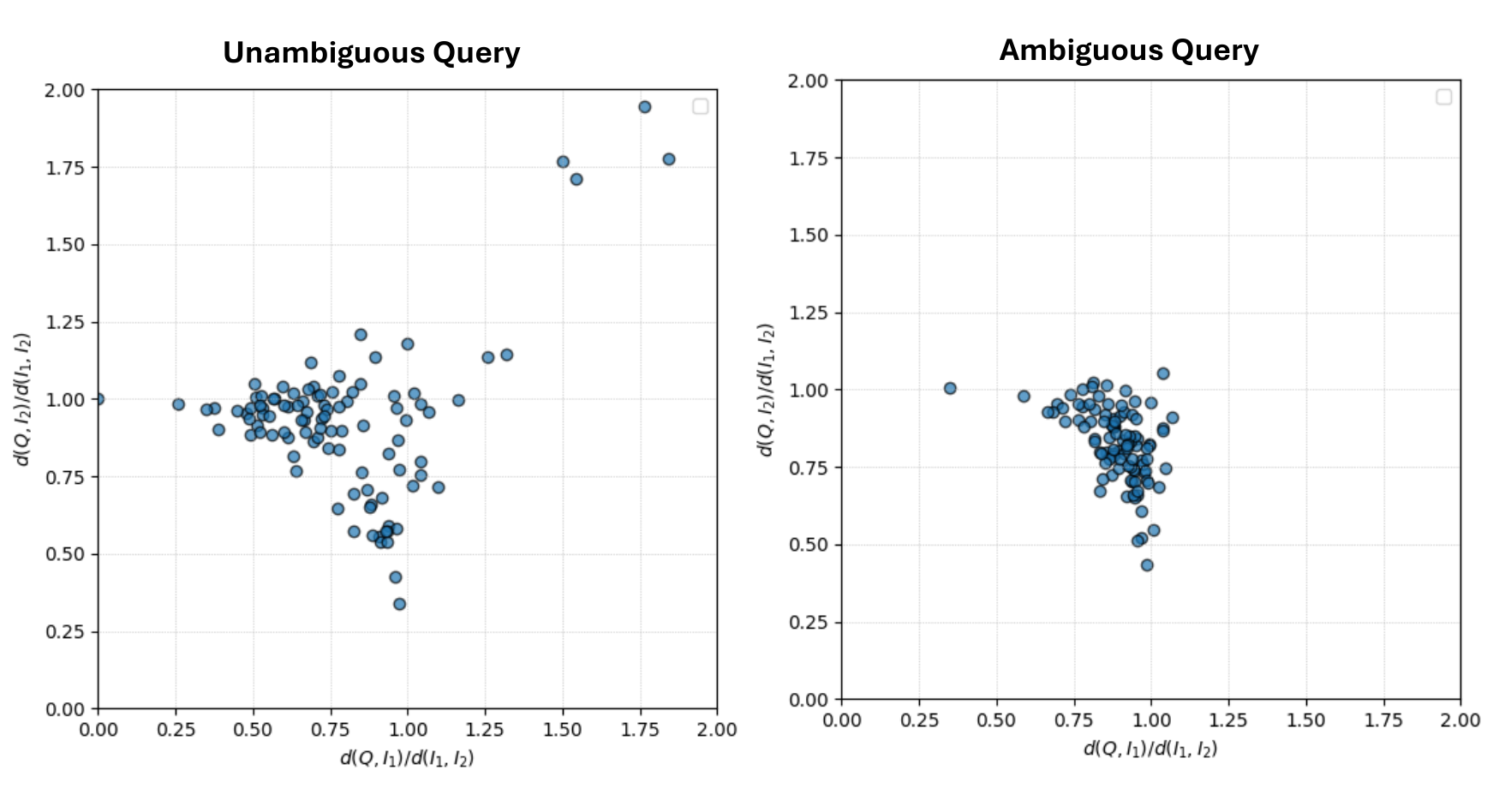}
     \caption{Normalized distances (Equation \ref{distance_calculation_2}) for the distances between questions and their interpretations using the ratios $\mathrm{D}_2(q, i_1) / \mathrm{D}_2(i_1, i_2)$ and $\mathrm{D}_2(q, i_2) / \mathrm{D}_2(i_1, i_2)$, compared to unambiguous and ambiguous questions' distance triplets cluster to the center of the map.}
    \label{fig:distanse_relationship}
\end{figure}

In Figure~\ref{fig:distanse_relationship}, we normalize the distances (Equation \ref{distance_calculation_2}) for the distances between $q, i_1, i_2$ using the ratios $\mathrm{D}_2(q, i_1) / \mathrm{D}_2(i_1, i_2)$ and $\mathrm{D}_2(q, i_2) / \mathrm{D}_2(i_1, i_2)$, and plot these normalized values to reveal potential patterns. In this case, no concept mask is applied during distance calculation. Our goal is to examine distance patterns when using equation~\ref{distance_calculation_2} with all activated concepts valid. 
For clarity, we visualize 100 samples for each case.
We find that compared to unambiguous questions, interpretations for ambiguous questions are more concentrated and more symmetrically distributed in their distances 
to the questions. 

\begin{figure}[h!]
\begin{subfigure}{\textwidth}
\centering
\includegraphics[width=0.75\linewidth]{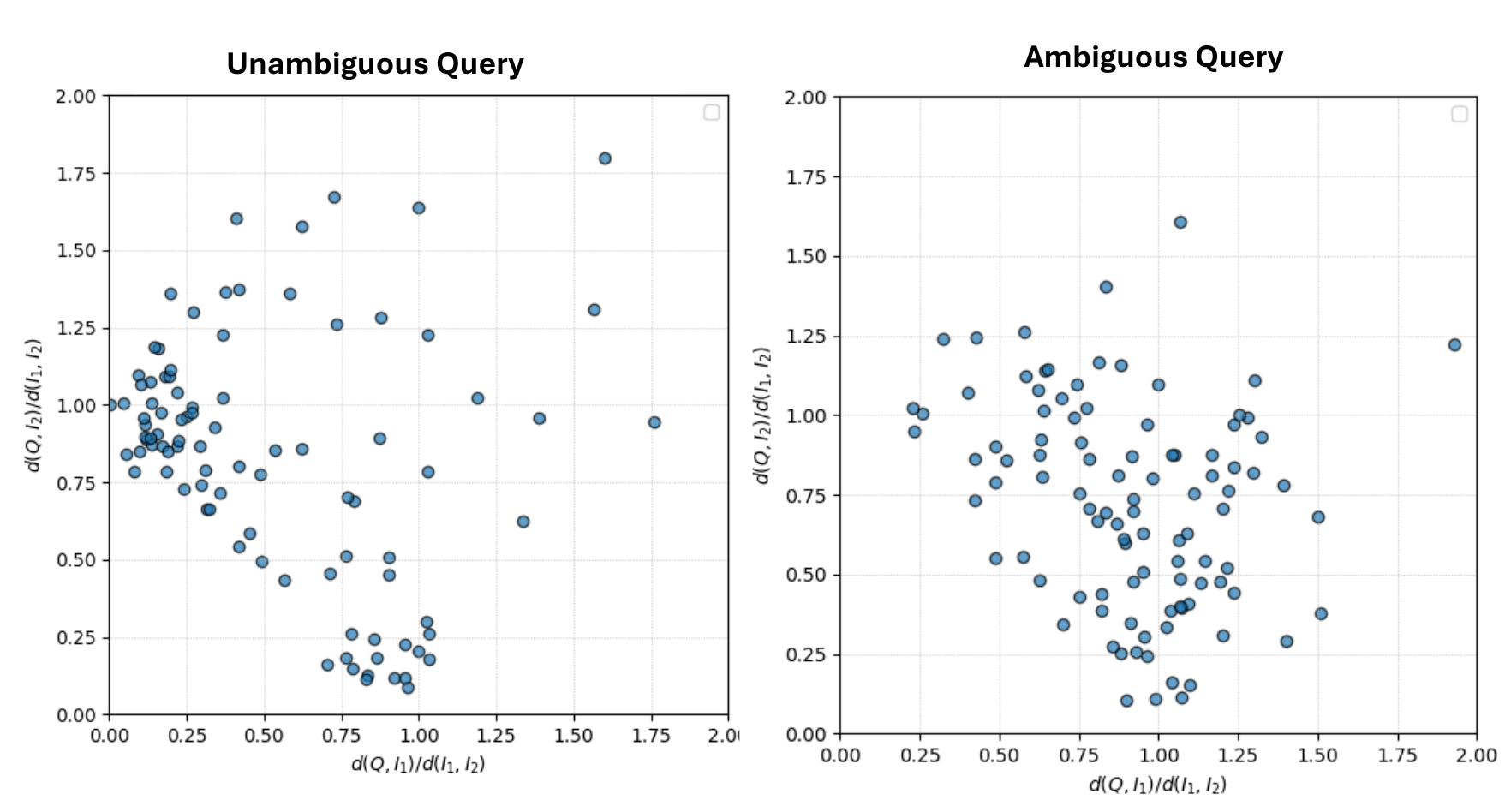}
\caption{Llama-3.3-70B-instruct model}
\label{fig:subim1}
\end{subfigure}
\begin{subfigure}{\textwidth}
\centering
\includegraphics[width=0.75\linewidth]{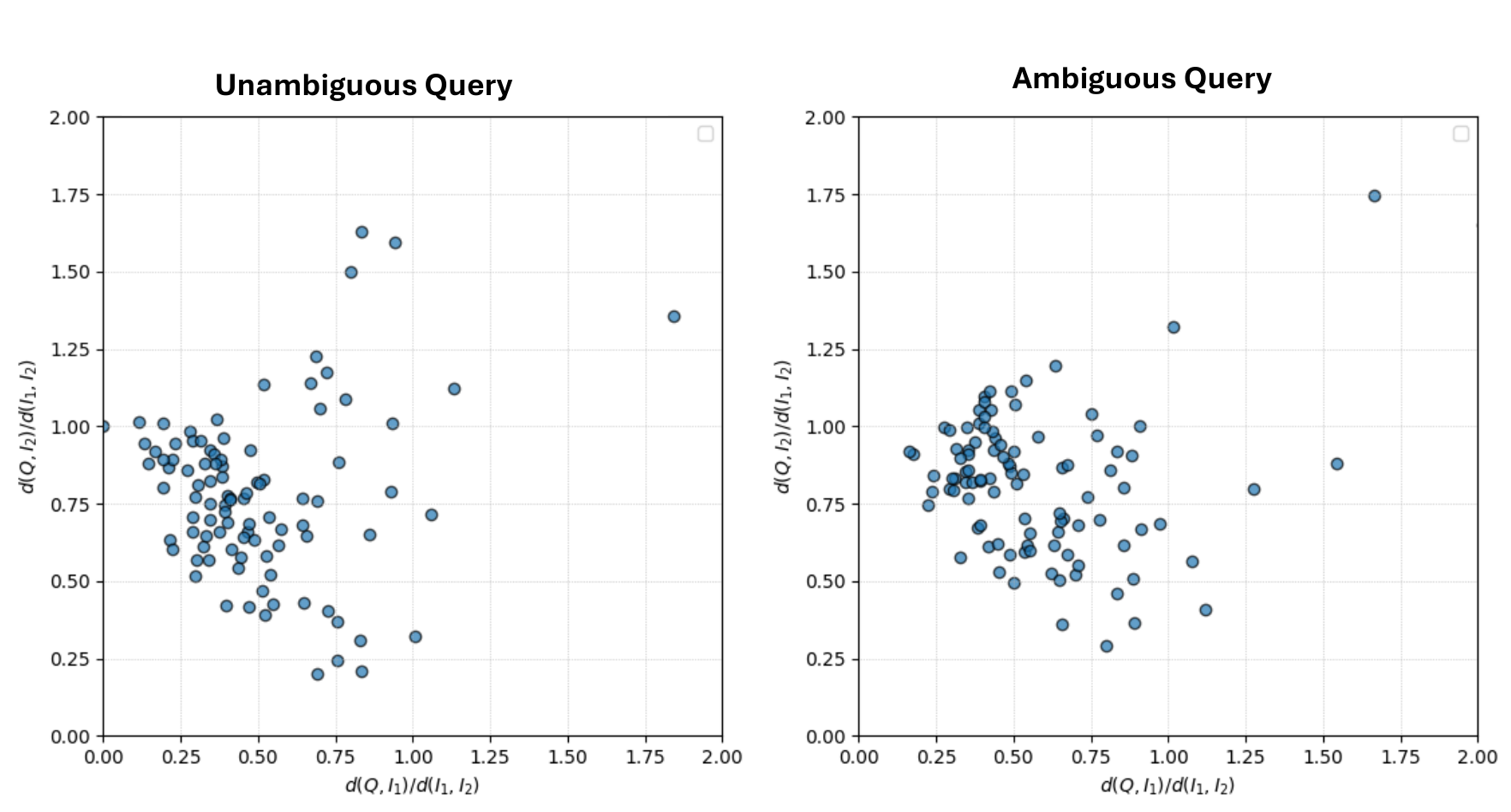}
\caption{SFR-Embedding-Mistral model}
\label{fig:subim2}
\end{subfigure}
\caption{Normalized distances for the baselines.}
\label{fig:Distance_relationship_baseline}
\end{figure}

Figure~\ref{fig:Distance_relationship_baseline} shows the results of the baselines, we can see that distance calculations with dense vectors generated by both generation and embedding models cannot show the symmetric pattern of ambiguous questions and their interpretations. As such, we can not distinguish each data point is ambiguous or not by their measurements.

\clearpage

\subsection{Ambiguity sensitivity improvement of LLMs with missing concept addition}\label{missing_concepts_adding_for_ambiguous_question}

To demonstrate the connection between targeted concepts and ambiguity resolution, we conducted experiments with LLaMA3-70B on questions involving scope ambiguity. Below is an example (an ambiguous question and its interpretations): \vspace{0.1cm}

\hspace{0.4cm}Ambiguous question: ”What brands of machinery are available in each machinery store?”;\\
\hspace*{0.4cm}Interpretation 1: ”Which brands of machinery are equally available in all machinery stores?”;\\
\hspace*{0.4cm}Interpretation 2: ”For each machinery store, show which brands of machinery are available.”\vspace{0.1cm}

As the missing concepts were found to correlate with tokens like “For,” “each,” “Which,” “What,” “all,” “?”, and “common”, we first identified the concepts invoked by these tokens, and then manually increased the activation values of these concepts to 1.0. As results in table \ref{tab:concept-activation} show, we found that the accuracy of ambiguity detection on unambiguous questions increased from 39.8\% to 60.5\%. In contrast, when the same number of random concepts were activated instead, the accuracy dropped to just 0.2\%. Although this will result in a 16.4\% decrease in the ambiguous question detection accuracy, the overall accuracy will increase from 46.31\% to 54.61\%. This indicates that only targeted concept activation helps identify ambiguity, whereas randomly activating concepts only bring interference.

\begin{table}[ht]
\centering
\caption{Effect of concept activation on ambiguity detection and overall accuracy.}
\label{tab:concept-activation}
\begin{tabular}{lcccc}
\toprule
\textbf{Metric} & \textbf{Baseline} & \textbf{Targeted activation} & \textbf{Random activation} & \textbf{vs. baseline} \\
\midrule
Unambiguous accuracy & 39.8\% & 60.5\% & 0.2\% & +20.7\,pp (targeted) \\
Ambiguous accuracy  & 59.3\%  & 42.8\%  & ---  & $-16.4$\,pp \\
Overall accuracy              & 46.31\% & 54.61\% & --- & +8.30\,pp \\
\bottomrule
\end{tabular}

\vspace{0.5ex}
\small \emph{Notes.} “pp” = percentage points. Unambiguous questions and ambiguous questions in dataset are unbalanced.
\end{table}

\section{Agentic Tool Calling: Prompt and Concept Matching Examples}
\subsection{Ambiguous Prompt Examples}\label{Ambiguity in agentic tool calling}

\begin{figure}[H]
    \centering
    \includegraphics[width=0.9\linewidth]{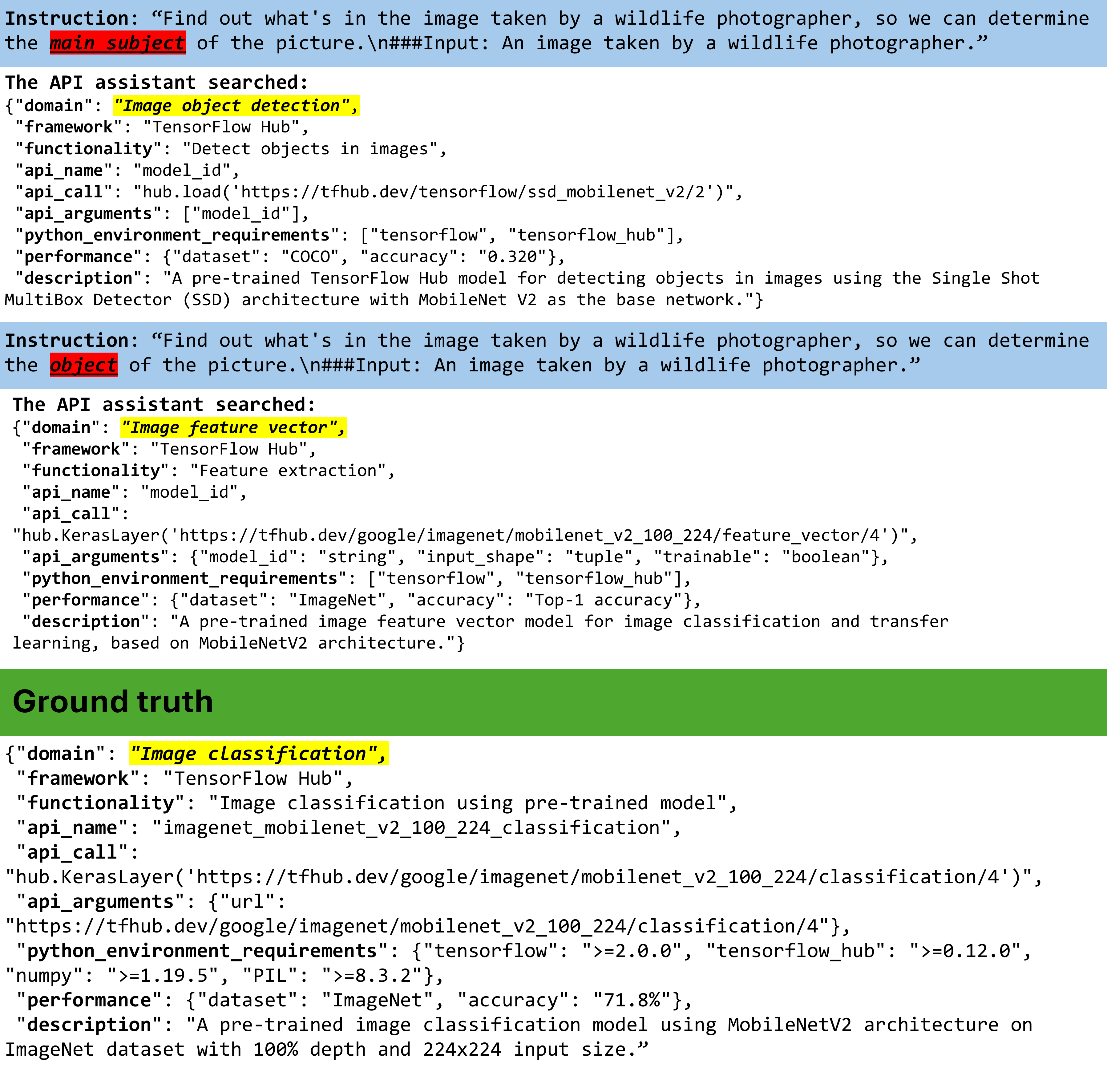}
     \caption{An example illustrating ambiguity in agentic tool calling (from the Gorilla dataset, using the Gorilla model as the API assistant). The red highlight marks differences in the instructions. Minor changes to the instruction can steer the LLM's answer, and may even shift the domain of the returned API.}
    \label{fig:API_ambiguty}
\end{figure}

\subsection{A Concept Matching Example}\label{Example for concepts mapping on agentic tool calling}

\begin{figure}[H]
    \centering
    \includegraphics[width=0.85\linewidth]{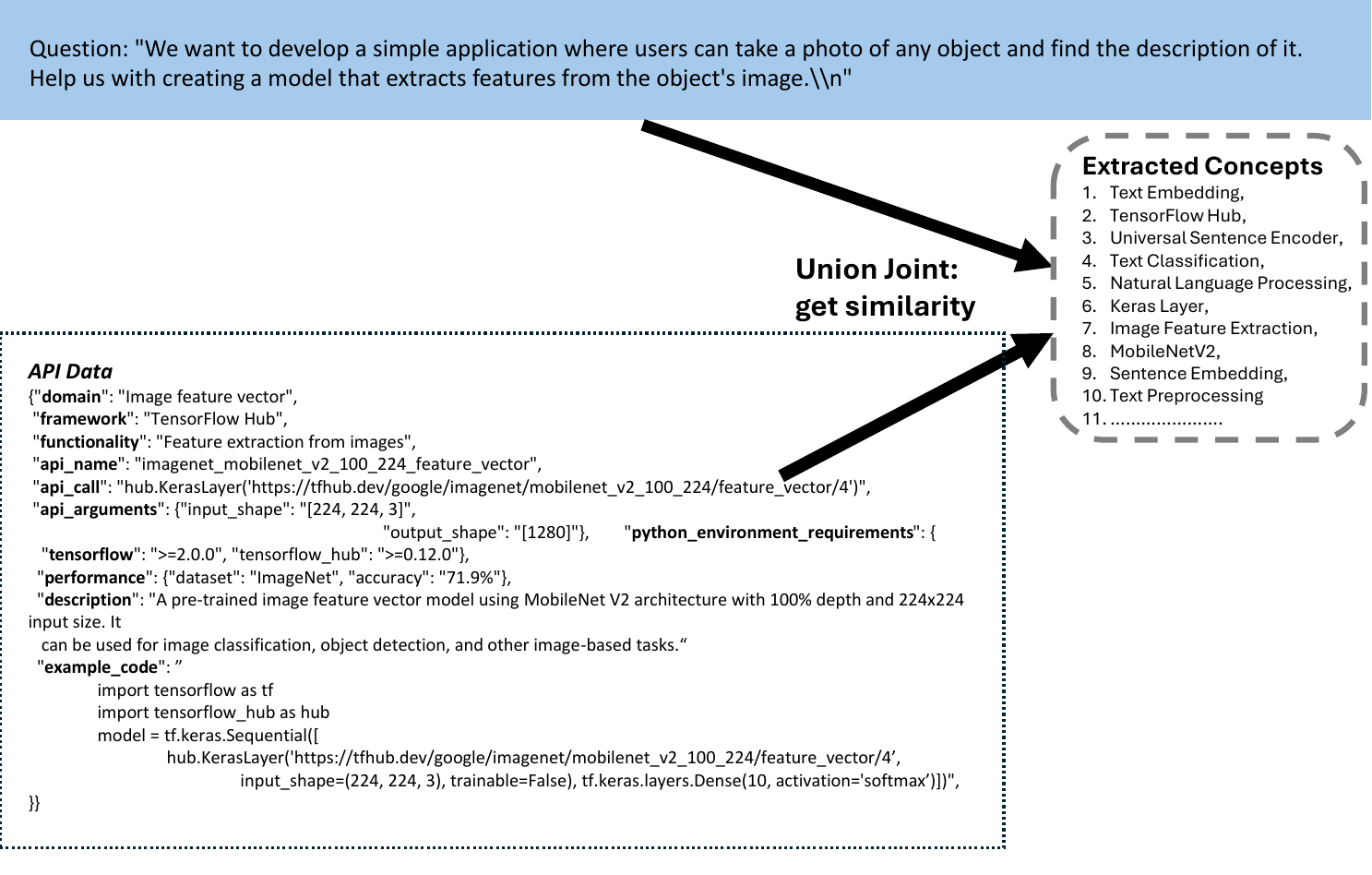}
     \caption{An example for getting similarity for extracted concepts by union joint.}
    \label{fig:Example_for_API_retrieval}
\end{figure}

Figure \ref{fig:Example_for_API_retrieval} illustrates how the concepts that involve which activated by the input question and which predicted by pre-trained predictor are matched to those in the structured API document through the union joint operator.

\section{Limitations}\label{Limitations}

Our method was evaluated on limited datasets. While results on both ambiguity and API datasets demonstrate its effectiveness, these datasets cover only a subset of known ambiguity scenarios, leaving it unclear whether our interpretation-generation method generalizes to other types of ambiguity in natural language. Investigating this question is left for future work.

\section{Computing Resources}\label{Computing Resources}

Our experiments were conducted on four NVIDIA H100 GPU node, each with 96GB memory.


\end{document}